%% file: eccv2020submission.tex
\documentclass[runningheads]{llncs}
\usepackage{graphicx}
\usepackage{comment}
\usepackage{amsmath,amssymb} 
\usepackage{color}

\input{sources/ourcommands}

\usepackage{epsfig}
\usepackage{subfig}
\DeclareMathOperator*{\argmax}{arg\,max}

\begin{document}
\pagestyle{headings}
\mainmatter
\def\ECCVSubNumber{6091}  

\title{DA4AD: End-to-End Deep Attention-based Visual Localization for Autonomous Driving} 

\titlerunning{ECCV-20 submission ID \ECCVSubNumber} 
\authorrunning{ECCV-20 submission ID \ECCVSubNumber} 
\author{Anonymous ECCV submission}
\institute{Paper ID \ECCVSubNumber}

\titlerunning{DA4AD: End-to-End Attention-based Visual Localization}
\author{Yao Zhou \hspace{0.4cm} Guowei Wan \hspace{0.4cm} Shenhua Hou \hspace{0.4cm} Li Yu \hspace{0.4cm} Gang Wang \\
	Xiaofei Rui \hspace{0.3cm} Shiyu Song\thanks{Author to whom correspondence should be addressed}}
\authorrunning{Y. Zhou, G. Wan, S. Hou, L. Yu, G. Wang, X. Rui, S. Song}
\institute{Baidu Autonomous Driving Technology Department (ADT)\\
	\email{\{zhouyao, wanguowei, houshenhua, yuli01, wanggang29,\\ruixiaofei, songshiyu\}@baidu.com}
}

\maketitle

\input{sources/abstract}
\input{sources/intro}
\input{sources/related}
\input{sources/problem}
\input{sources/method}
\input{sources/exp}
\input{sources/conclusion}

%
%
\bibliographystyle{splncs04}
\bibliography{egbib}
\end{document}

%% file: sources/ourcommands.tex
\usepackage{color}
\definecolor{green}{rgb}{0, 0.5, 0}
\definecolor{orange}{rgb}{0.8, 0.6, 0.2}
\definecolor{red}{rgb}{1.0, 0.0, 0.0}
\definecolor{teal}{rgb}{0.0, 0.4, 0.4}
\definecolor{purple}{rgb}{0.65,0,0.65}
\definecolor{saffron}{rgb}{0.95,0.75,0.2}
\definecolor{turquoise}{rgb}{0.0,0.5,0.5}
\definecolor {mygray}{gray}{.9}

\newcommand{\shiyu}[1]{{\color{black}#1}}

\usepackage{booktabs}
\usepackage{colortbl}
\usepackage{makecell}
\usepackage{multirow}
\usepackage{multicol}
\usepackage{array}
\usepackage{gensymb}
\newcolumntype{L}[1]{>{\raggedright\arraybackslash}p{#1}}
\newcolumntype{C}[1]{>{\centering\arraybackslash}p{#1}}
\newcolumntype{R}[1]{>{\raggedleft\arraybackslash}p{#1}}
\usepackage{amsmath}

\usepackage{tabu}
\usepackage{graphicx}

%% file: sources/abstract.tex
\begin{abstract}
We present a visual localization framework based on novel deep attention aware features for autonomous driving that achieves centimeter level localization accuracy. Conventional approaches to the visual localization problem rely on handcrafted features or human-made objects on the road. They are known to be either prone to unstable matching caused by severe appearance or lighting changes, or too scarce to deliver constant and robust localization results in challenging scenarios. In this work, we seek to exploit the deep attention mechanism to search for salient, distinctive and stable features that are good for long-term matching in the scene through a novel end-to-end deep neural network. Furthermore, our learned feature descriptors are demonstrated to be competent to establish robust matches and therefore successfully estimate the optimal camera poses with high precision. We comprehensively validate the effectiveness of our method using a freshly collected dataset with high-quality ground truth trajectories and hardware synchronization between sensors. Results demonstrate that our method achieves a competitive localization accuracy when compared to the LiDAR-based localization solutions under various challenging circumstances, leading to a potential low-cost localization solution for autonomous driving.
\end{abstract}

%% file: sources/intro.tex
\section{Introduction}
\label{section:intro}

Localization is a fundamental task in a self-driving car system.
To exploit high definition (HD) maps as priors for robust perception and safe motion planning, this requires the localization system to reach centimeter-level accuracy \cite{barsan2018learning}.

\begin{figure}[!!t]
	\centering
	\includegraphics[width=1.0\linewidth]{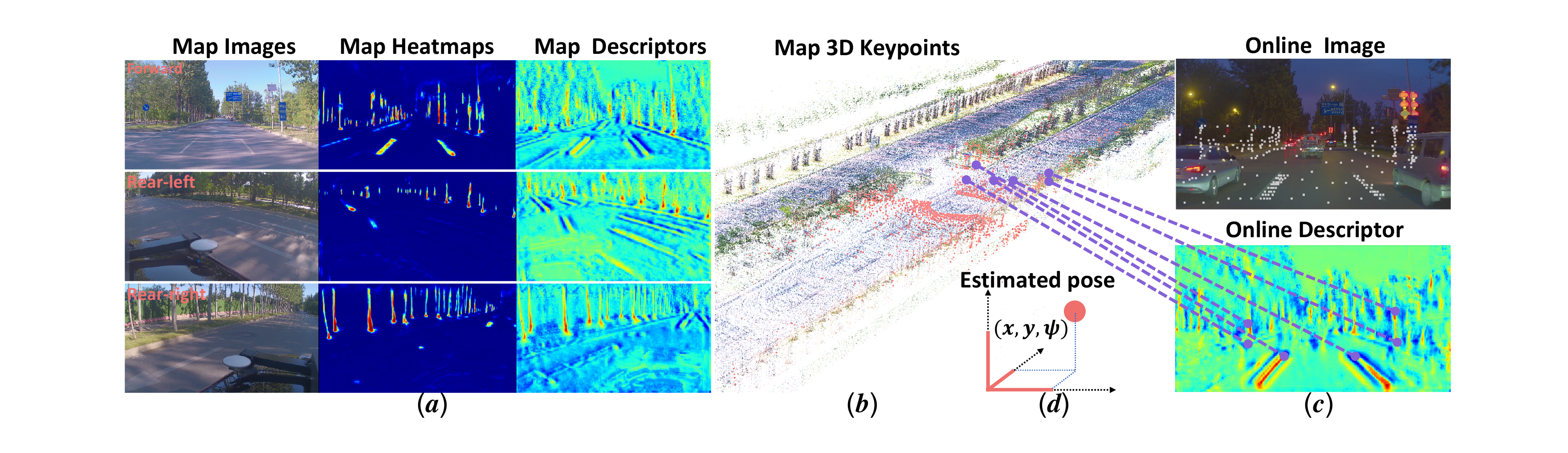}
	\caption{
		The major steps of our proposed framework:
			(a) The heatmaps (middle) and descriptor maps (right) extracted by the local feature embedding module.
		    (b) Map 3D keypoints are selected by the attentive keypoint selection module in accordance with the map heatmaps.
	        (c) The neighboring keypoints in the map are projected onto the online image (top) given a set of candidate poses. The corresponding feature descriptors in the online image are found.
            (d) The optimal camera pose is estimated by evaluating the overall feature matching cost.
	}
	\label{fig:intro}
\end{figure}

Despite many decades of research, building a long-term, precise and reliable localization system using low-cost sensors, such as automotive and consumer-grade GPS/IMU and cameras, is still an open-ended and challenging problem.
\shiyu{Compared to the LiDAR, cameras are passive sensors meaning that they are more susceptible to appearance changes caused by varying lighting conditions or changes in viewpoint.}
It is known that handcrafted point features (DIRD \cite{lategahn2013learn,lategahn2014vision}, FREAK \cite{alahi2012freak,burki2019vizard}, BRIEF \cite{calonder2010brief,linegar2015work} et al.) suffer from unreliable feature matching under large lighting or viewpoint change, leading to localization failure.
Even when using recent deep features \cite{yi2016eccvlift,savinov2017quad,he2018localdes,deTone2018superpoint}, local 3D-2D matching is prone to fail under strong visual changes in practice due to the lack of repeatability in the keypoint detector \cite{sattler2018benchmarking,sarlin2019coarse,germain2019sparse}.
Another alternative to these methods is to leverage human-made objects, which encode appearance and semantics in the scene, such as lane \cite{schreiber2013laneloc,cui2014real} or sign \cite{ranganathan2013light} markings on the road \cite{jo2015precise,suhr2016sensor}, road curbs, poles \cite{yu2014monocular} and so on.
Those features are typically considered relatively stable and can be easily recognized as they are built by humans for specific purposes and also used by human drivers to aid their driving behavior.
Nevertheless, those methods are only good for environments with rich human-made features but easily fail in challenging scenarios that lack them, for example, road sections with worn-out markings under poor maintenance, rural streets with no lane markings or other open spaces without clear signs.
Furthermore, these carefully selected semantic signs or markings typically only cover a small area in an image.
One obvious design paradox in a mainstream visual localization system is that it suffers from the absence of distinctive features, however, at the same time, it deliberately abandons rich and important information in an image by solely relying on these human-made features.


In this work, titled ``DA4AD'' (deep attention for autonomous driving), we address the aforementioned problems by building a visual localization system that trains a novel end-to-end deep neural network (DNN) to extract learning-based feature descriptors, select attentive keypoints from map images, match them against online images and infer optimal poses through a differentiable cost volume.
Inspired by prior works \cite{noh2017large,Dusmanu_2019_CVPR} that utilize the attention mechanism, our intuition is that we seek to effectively select a subset of the points in the map images as attentive keypoints. They are stable features in the scene and good for long-term matching.
To this end, we first construct an image pyramid and train fully convolutional networks (FCN) to extract dense features from different scales on them independently.
Using shared backbone networks, dense heatmaps from different scales are simultaneously estimated to explicitly evaluate the attention scores of these features for their capabilities in conducting robust feature matching under strong visual changes.
To build a prior map, we process our map images and store the selected attentive keypoints, extracted features, and 3D coordinates into a database.
The 3D coordinates are obtained from LiDAR scans which are only used for mapping.
During the online localization stage, given a predicted prior vehicle pose as input, we query the nearest neighboring map image with selected keypoints in it from the database.
We then sample a set of candidate poses around the prior pose.
By projecting the 3D map keypoints onto the online image using each candidate pose, the matched 2D points in the online image can be found and their local features have been extracted accordingly.
Finally, given these local feature descriptor pairs from both the online and map image as input, by evaluating overall feature matching cost across all the candidate poses, the optimal estimation can be obtained.
More importantly, in this final feature matching step, we infer the unknown camera poses through a differentiable multi-dimensional matching cost volume in the solution space, yielding a trainable end-to-end architecture.
Compared to other works \cite{yi2016eccvlift,savinov2017quad,he2018localdes,deTone2018superpoint,Dusmanu_2019_CVPR} that learn deep feature descriptors, this architecture allows our feature representation and attention score estimation function to be trained jointly by backpropagation and optimized towards our eventual goal that is to minimize the absolute localization error.
Furthermore, it bypasses the repeatability crisis in keypoint detectors in an efficient way.
This end-to-end architecture design is the key to boost the overall performance of the system.

To summarize, our main contributions are:
\begin{itemize}
	\item A novel visual localization framework for autonomous driving, yielding centimeter level precision under various challenging lighting conditions.
	\item Use of the attention mechanism and deep features through a novel end-to-end DNN which is the key to \shiyu{boost performance.}
	\item Rigorous tests and benchmarks against several methods using a new dataset with high-quality ground truth trajectories and hardware camera, LiDAR, IMU timestamp synchronization. We shall release the dataset shortly.
\end{itemize}

%% file: sources/related.tex
\section{Related Work}
\label{section:related}

In the recent two decades, there has been a breakthrough in LiDAR-based localization technologies that has led to compelling performance
\cite{Levinson_etal2007,levinson2010robust,wolcott2015fast,Wolcott_etal2017,wan2018robust,barsan2018learning,lu2019l3}. However, 
camera-based solutions are particularly favored by car manufacturers and Tier 1 suppliers due to their low cost.

\textbf{Structure based}
One important category of methods utilizes human-made structures in the environment.
M. Schreiber et al. \cite{schreiber2013laneloc} localize the vehicle using a stereo camera and by matching curbs and lane segments in a map.
D. Cui et al. \cite{cui2014real,cui2015real} conversely detect consecutive lanes instead of lane segments and globally locate them by registering the lane shape with the map.
Furthermore, K. Jo et al. \cite{jo2015precise} introduce an around-view monitoring (AVM) system and improve the localization performance by benefiting from the detected lane markings within it.
A. Ranganathan et al. \cite{ranganathan2013light,wu2013vehicle} propose to use signs marked on the road instead of the lane markings to localize the vehicles. 
J. K. Suhr et al. \cite{suhr2016sensor} further build a localization system using lane contours and road signs expressed by a minimum number of points solved in a particle filter framework.
Y. Yu et al. \cite{yu2014monocular} utilize line segments in the scene, such as lane markings, poles, building edges together with sparse feature points, and define a road structural feature (RSF) for localization.

\textbf{Low-level Feature based}
Another category of methods employs low-level features.
H. Lategahn et al. \cite{lategahn2013urban,lategahn2014vision} detect salient 2D sparse points in the online image and matched with the queried 3D landmarks from the map using handcrafted DIRD descriptors \cite{lategahn2013learn}, and the 6 DoF poses are solved in a probabilistic factor graph.
Since the number of successfully matched landmarks is crucial to the performance, H. Lategahn et al. \cite{linegar2015work} later propose to learn to select the most relevant landmarks to improve the overall long-term localization performance.
Moving forward, \cite{linegar2016made} introduces linear SVM classifiers to recognize distinctive landmarks in the environment through a unsupervised mining procedure.
Most recently, M. B{\"u}rki et al. \cite{burki2019vizard} build a vision-based localization system with a classical 2D-3D correspondence detection-query-matching pipeline leveraging the FREAK \cite{alahi2012freak} descriptor.
Similar to us, LAPS-II \cite{stewart2012laps,maddern2014laps} also utilizes the 3D structure and performs 6 DoF visual localization by first transforming RGB images to an illumination invariant color space and then finding the optimal vehicle poses by minimizing the normalized information distance (NID), which can still be considered as a handcrafted way to embed feature representations.
Furthermore, \cite{wolcott2014visual,pascoe2015direct,neubert2017sampling,chen2019enforcenet} propose to localize their vehicles by generating numerous synthetic views of the environment rendered with LiDAR intensity, depth or RGB values, and comparing them to the camera images to find the optimal candidate by minimizing a handcrafted or learned cost.
Recently, deep features \cite{yi2016eccvlift,savinov2017quad,he2018localdes,deTone2018superpoint,Dusmanu_2019_CVPR,von2020gn} have been proposed to replace these traditional handcrafted ones.
\shiyu{GN-Net \cite{von2020gn} proposes to train deep features in a self-supervised manner (similar to us) with the newly proposed Gauss-Newton loss.}
The work most related to our approach is M. Dusmanu et al. \cite{Dusmanu_2019_CVPR}. It proposes a similar attentive \textit{describe-and-detect} approach to us, but fails to integrate the feature descriptor and detector training process into a specific application task through an end-to-end DNN, which is the key to boost the performance.


Other recent attempts include DeLS-3D \cite{wang2018dels} that proposes to directly estimate the camera pose by comparing the online image with a synthetic view rendered with semantic labels given a predicted prior pose, and T. Caselitz \cite{caselitz2016monocular} that localizes its self-position by matching reconstructed semi-dense 3D points from image features against a 3D LiDAR map.
\shiyu{Similar to us, H. Germain et al. \cite{germain2019sparse} proposes to only detect (``select'' in our work) features in the reference image to bypass the repeatability crisis in the keypoint detection.}
A large body of literature \cite{Cipolla2015,kendall2017geometric,naseer2017loc,walch2017image,brahmbhatt2018geometry,taira2018inloc,toft2018semantic,valada2018loc,radwan2018vlocnet,sarlin2019coarse,sattler2019aprrpr} focuses on solving vision-based localization problems for other applications instead of the autonomous driving, which are not discussed in detail here.

%% file: sources/problem.tex
\section{Problem Statement}
\label{section:problem}

Our problem definition is similar to previous works \cite{wang2018dels,lu2019l3}, where the input to the localization system involves a pre-built map that encodes the memory of the environment in history, and a predicted coarse camera pose usually estimated by accumulating the localization estimation of the previous frame with the incremental motion estimation obtained from an IMU sensor.
At the system initialization stage, this prior pose can be obtained using GPS, other image retrieval techniques or Wi-Fi fingerprinting. Our map representation contains 2D point features together with global 3D coordinates.

Therefore, given an online image, our task is to seek an optimal offset between the final and predicted pose by 3D-2D matching the features from the pre-built map to the ones from the online image.
For better efficiency and robustness, we follow localization systems \cite{levinson2010robust,wan2018robust,lu2019l3} for autonomous driving, and only the 2D horizontal and heading offset $(\Delta x, \Delta y, \Delta \psi)$ is estimated.

%% file: sources/method.tex
\section{Method}
\label{section:method}

\begin{figure*}[!htbp]
	\centering
	\includegraphics[width=\linewidth]{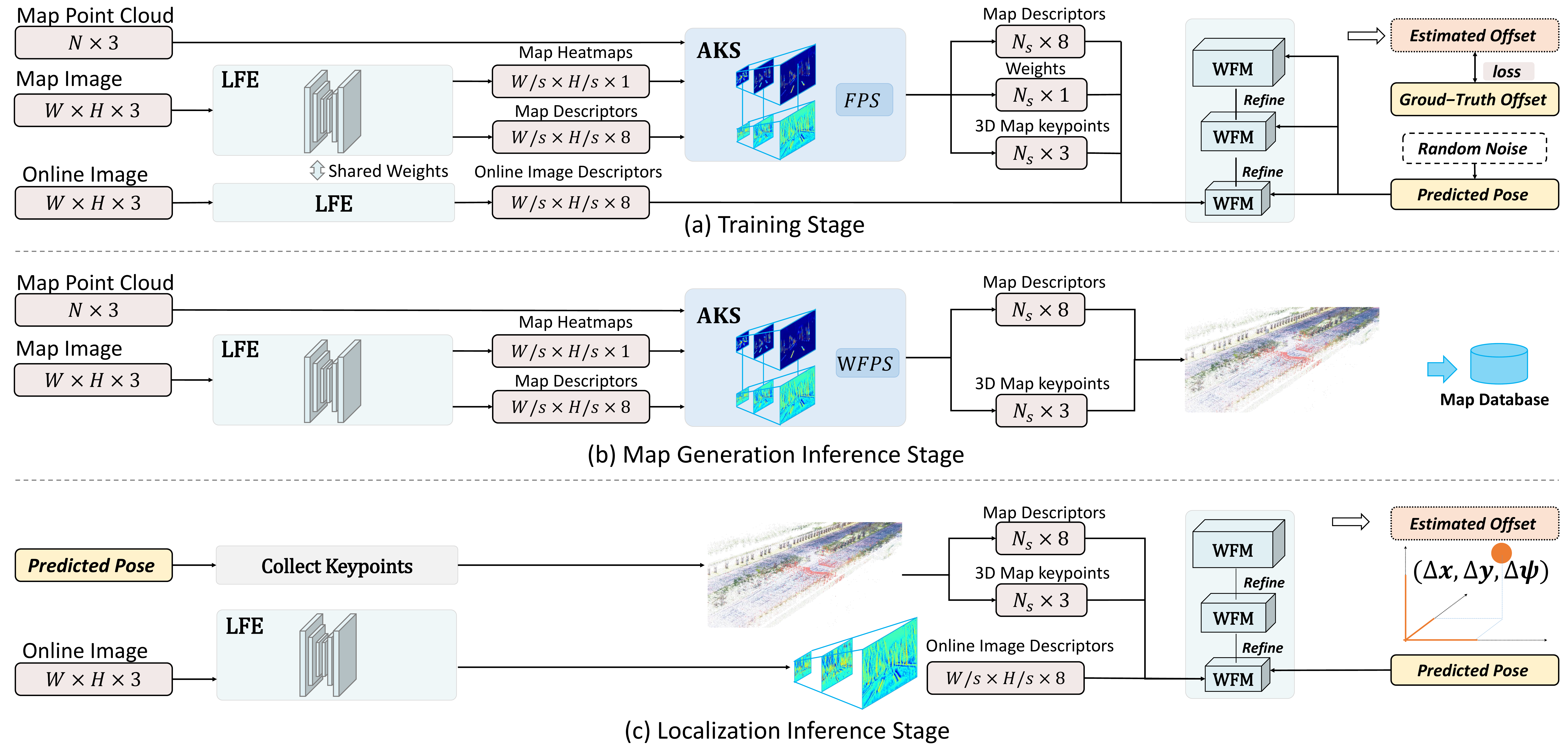}
	\caption{
		The network architecture and system workflow of the proposed vision-based localization framework \shiyu{based on} end-to-end deep attention aware features in three different stages: a) training; b) map generation; c) online localization.
	}
	\label{fig:architecture}
\end{figure*}

There are three different stages in the system: (i) network training; (ii) map generation; (iii) online localization.
Both the map generation and online localization can be considered as inferences of the trained network.
The network architecture of the proposed framework in different stages is shown in Figure~\ref{fig:architecture}.

\subsection{Network Architecture}

The overall architecture can be decomposed into three main modules: (i) local feature embedding (LFE); (ii) attentive keypoint selection (AKS); (iii) weighted feature matching (WFM).
To seek the best performance, we may choose to use different algorithms or strategies in the same module when the system is in different stages.
These choices are introduced in detail in Section~\ref{subsec:backbone}, \ref{subsec:keypoints} and \ref{subsec:l3net}. \shiyu{The effectiveness of them is validated thoroughly in our experiments.}

\textbf{LFE module}
We seek to extract good local feature descriptors together with their relevance weights (attention scores) to our task represented as a heatmap in an image.
Ideally, these extracted descriptors should be robust for matching under appearance changes caused by different lighting conditions or seasons.
The attention scores should highlight reliable objects and avoid interferences and noises in the scene.

\textbf{AKS module}
\shiyu{Despite the fact that our LFE module extracts dense features,} similar to \cite{Dusmanu_2019_CVPR}, we adopt a \textit{describe-and-select} approach to select a set of keypoints that are good for long-term matching and save them in the map database.

\textbf{WFM module}
Given 3D keypoints associated with their 2D feature descriptors from the map images and dense features extracted from the online image, the WFM module estimates the optimal pose by sampling a set of candidate poses around the prior pose and evaluating the matching cost given each candidate.

\subsection{System Workflow}
\label{subsec:workflow}


\textbf{Training}
The training stage involves all the three modules, LFE, AKS, and WFM.
\shiyu{First of all, given a predicted pose, its closest map image in the Euclidean distance is selected. Next, the LFE module extracts the dense features from both the online and map images, and the attention heatmap from the map image accordingly.
Based on the attention scores from the heatmap, the AKS module selects good features from the map image as the keypoints.
Then their associated 3D coordinates are obtained from LiDAR point cloud projections.
Finally, given these 3D keypoints and feature descriptors as input, the WFM module seeks to find the optimal pose offset by searching in a 3D cost volume, and the optimal pose offset is compared with the ground truth pose to produce the training loss.}

\textbf{Map Generation}
After training, there is a designated map generation step using a sub-portion of the network as shown in Figure~\ref{fig:architecture}.
To build the map and test the system, we did multiple trials of data collection on the same road.
One of them is used for mapping.
Given the LiDAR scans and the ground truth vehicle poses (see Section~\ref{section:dataset} for details), the global 3D coordinates of LiDAR points can be obtained readily.
Note that the LiDAR sensors and ground truth poses are used for mapping purposes only.
First, the map image pixels are associated with global 3D coordinates by projecting 3D LiDAR points onto the image, given the ground truth vehicle poses.
Attention heatmaps and feature maps of different resolutions in the map image are then estimated by the LFE network inference.
Next, a set of keypoints are selected for different resolutions in the pyramid in the AKS module.
As a whole, we save keypoints together with their D-dimensional descriptors and 3D coordinates in the map database.

\textbf{Online Localization}
During the localization stage, feature maps of different resolutions in the online image are again estimated by the LFE network inference.
We collect the keypoints with their feature descriptors and global 3D coordinates from the nearest map image given the predicted camera pose.
Then these keypoints are projected onto the online image given the sampled candidate poses in the cost volume we built in the WFM module.
Three feature matching networks of different resolutions cascade to achieve a coarse-to-fine inference and output the estimated vehicle pose.

\begin{figure*}[!htbp]
	\centering
	\includegraphics[width=\linewidth]{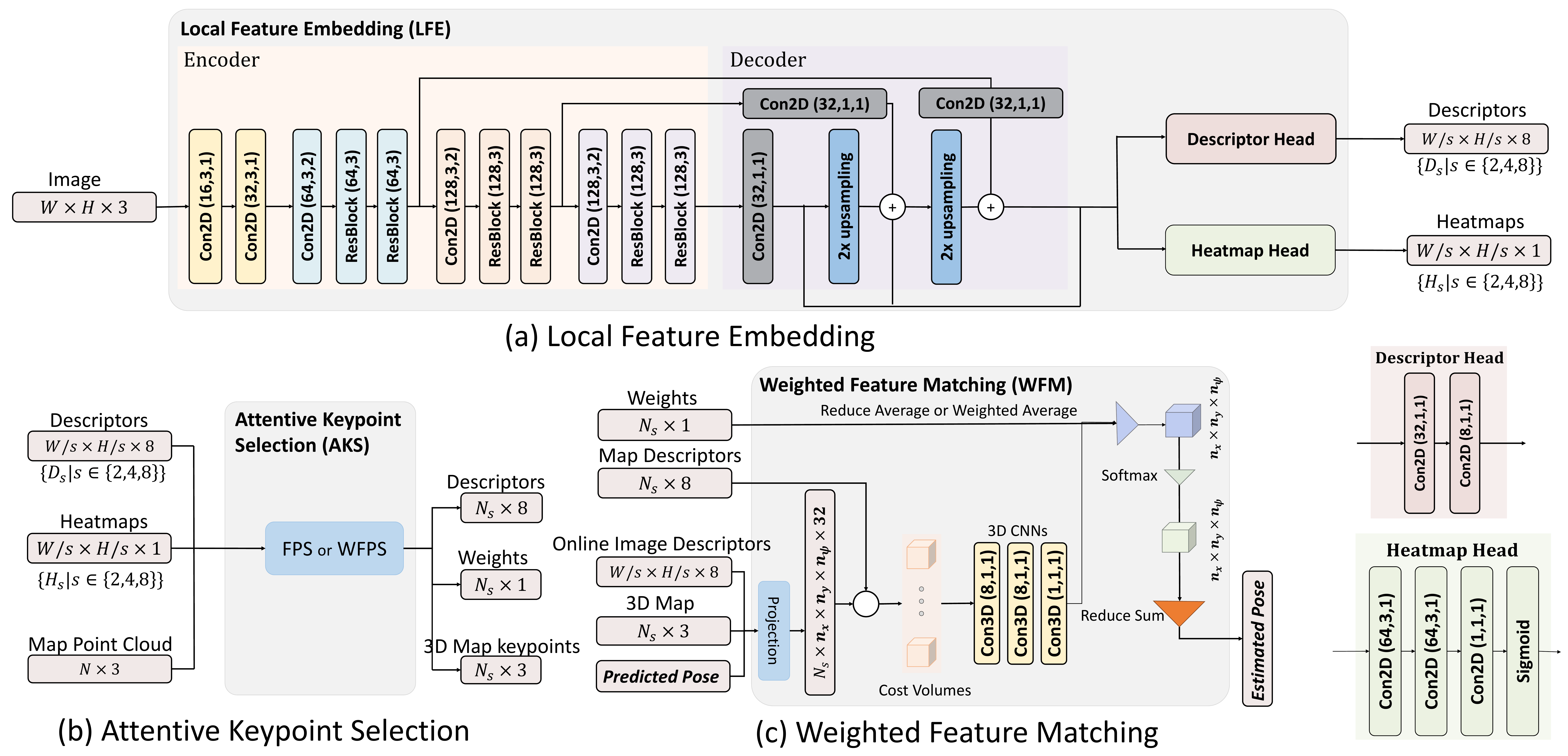}
	\caption{
		The illustration of the network structure of the three main modules: (a) local feature embedding (LFE); (b) attentive keypoint selection (AKS); (c) weighted feature matching (WFM).
	}
	\label{fig:subnetwork}
\end{figure*}

\subsection{Local Feature Embedding}
\label{subsec:backbone}

The same LFE module is used in all three different stages.
We employ a network architecture similar to the feature pyramid network (FPN) introduced by T. Lin et al. \cite{lin2017feature} as shown in Figure~\ref{fig:subnetwork}(a).
With the lateral connections merging feature maps of the same spatial size from the encoder path to the decoder, the FPN can enhance high-level semantic features at all scales, thus harvesting a more powerful feature extractor.
In our encoder, we have an FPN consisting of 17 layers that can be decomposed into four stages.
The first stage consists of two 2D convolutional layers where the numbers in brackets are channel, kernel and stride sizes, respectively.
Starting from the second stage, each stage consists of a 2D convolutional layer with stride size 2 and two residual blocks introduced in the ResNet \cite{he2016deep}.
Each residual block is composed of two $3 \times 3$ convolutional layers.
In the decoder, after a 2D convolutional layer, upsampling layers are applied to hallucinate higher resolution features from coarser but semantically stronger features.
Features of the same resolution from the encoder are merged to enhance these features in the decoder through the aforementioned lateral connections that element-wise average them.
The outputs of the decoder are feature maps with different resolutions of the original image.
They are fed into two different network heads as shown in the bottom right of Figure~\ref{fig:subnetwork}, that are responsible for descriptor extraction and attention heatmap estimation, respectively.
The feature descriptors are represented as D-dimensional vectors that are competent for robust matching under severe appearance changes caused by varying lighting or viewpoint conditions.
The heatmap is composed of [$0$ - $1$] scalars that are used as relevance weights in our attention-based keypoint selection and feature matching modules in Section~\ref{subsec:keypoints} and \ref{subsec:l3net}.
To be more specific, our descriptor map output is a 3D tensor $F \in \mathbb{R}^{\frac{H}{s} \times \frac{W}{s} \times D}$, where $s \in {2, 4, 8}$ is the scale factor and $D = 8$ is the descriptor dimension size.
Our attention heatmap output is an image $W \in [0, 1]^{\frac{H}{s} \times \frac{W}{s}}$.


\subsection{Attentive Keypoint Selection}
\label{subsec:keypoints}

During the study, we learned that different keypoint selection strategies have a considerable impact on the overall performance of the system.
The AKS module is used in two stages: the training and map generation.
As we are solving a geometric problem, it's well known that a set of keypoints that are almost evenly distributed in the geometric space rather than clustered together are crucial.
We find that the proposed methods are superior to other more natural choices, for example top-K.

We consider two selection strategies, which are the farthest point sampling (FPS) algorithm \cite{eldar1997farthest} and a variant of it, the weighted FPS (WFPS) algorithm as shown in Figure~\ref{fig:subnetwork}(b).
Given a set of selected points $S$ and unselected $Q$, if we seek to iteratively select a new point $\hat{q}$ from $Q$, the FPS algorithm calculates
\begin{small}
	\begin{equation}
	\hat{q} = \argmax_{q \in Q}(\min_{s \in S}(d(q, s))).
	\end{equation}
\end{small}
In our WFPS algorithm, we instead calculate
\begin{small}
	\begin{equation}
	\hat{q} = \argmax_{q \in Q}(w(q)\min_{s \in S}(d(q, s))),
	\end{equation}
\end{small}
where $w(q)$ is the relevance attention weight of the query point $q$.

During the training stage, we aim to uniformly learn the attention scores of all the candidates, therefore it's necessary to have an efficient stochastic selection strategy.
To this end, first of all, $K$ candidate points are randomly selected and sent to the GPU cache.
Next, we apply the FPS algorithm to select the keypoints among them.
\shiyu{Interestingly, we find that a WFPS keypoint selection algorithm + a \textit{weighted average} marginalization operation (introduced in Section~\ref{subsec:l3net}) leads to catastrophic models.}
Our insight is that \textit{weighted average} contributes as a positive feedback to the WFPS keypoint selection policy.
A few good keypoints gain high weights rapidly, stopping others from being selected during the training, leading to a heatmap in where there are only a few clustered high weight points and of which the remaining part is simply empty.
A WFPS keypoint + a \textit{reduce average} marginalization operation (introduced in Section~\ref{subsec:l3net}) is not a valid approach as the heatmap network can not be trained without effective incorporation of the attention weights.

During the map generation stage, we need an algorithm that can select good keypoints by effectively incorporating the trained attention weights.
For this reason, again we first randomly selected $K$ candidate points and then stick to the WFPS during map generation considering it as a density sampling function with the heatmap as probabilities.

In order to associate the 2D feature descriptors with 3D coordinates, we project 3D LiDAR points onto the image.
Given the fact that not all image pixels are associated with LiDAR points, only the sparse 2D pixels with known 3D coordinates are considered as candidates, from which we select keypoints that are good for matching.
Please refer to the supplemental materials for exact numbers of keypoints for different resolutions.

\subsection{Weighted Feature Matching}
\label{subsec:l3net}
Traditional approaches typically utilize a PnP solver \cite{haralick1994pnp} within a RANSAC \cite{fischler1981ransac} framework to solve the camera pose estimation problem given a set of 2D-3D correspondings.
Unfortunately, these matching approaches including the outlier rejection step, are non-differentiable and thus prevent them from the feature and attention learning through backpropagation during the training stage.
$L^3$-Net \cite{lu2019l3} introduced a feature matching and pose estimation method that leverages a differentiable 3D cost volume to evaluate the matching cost given the pose offset and the corresponding feature descriptor pairs from the online and map images.

In the following, we improve the original $L^3$-Net design by coming up with solutions to incorporate attention weights and making them effectively trainable.
The network architecture is illustrated in Figure~\ref{fig:subnetwork}(c).

\textbf{Cost Volume}
Similar to the implementation of $L^3$-Net \cite{lu2019l3}, we build a cost volume $N_s \times n_x \times n_y \times n_\psi$, where $N_s$ is the number of selected keypoints, $n_x$, $n_y$ and $n_\psi$ are the grid sizes in each dimension.
To be more specific, given the predicted pose as the cost volume center, we divide its adjacent space into a three-dimensional grid evenly, denoted as $\{\Delta T = (\Delta x_i, \Delta y_j, \Delta \psi_k) | 1 \leq i \leq n_x, 1 \leq j \leq n_y, 1 \leq k \leq n_\psi\}$.
Nodes in this cost volume are candidate poses from which we desire to evaluate their corresponding feature pairs and find the optimal solution.
By projecting the selected 3D keypoints in the map images onto the online images using each candidate pose, the corresponding local feature descriptors can be calculated by applying the bilinear interpolation on the descriptor map of the online image.
Unlike the implementation of $L^3$-Net where it computes the element-wise $L2$ distance between both descriptors from the online and map images, we calculate the total $L2$ distance between them, bringing a single-dimensional cost scalar.
The cost scalar is then processed by a three-layer 3D CNNs with a kernel of Conv3D(8,1,1)-Conv3D(8,1,1)-Conv3D(1,1,1), and the result is denoted as $P(p, \Delta T)$, where $p$ is a keypoint out of N.

\textbf{Marginalization}
In the original implementation of $L^3$-Net, the regularized matching cost volume $N_s \times n_x \times n_y \times n_\psi$ is marginalized into a $n_x \times n_y \times n_\psi$ one across the keypoint dimension by applying a \textit{reduce average} operation.
Following \cite{noh2017large}, \shiyu{how to effectively incorporate the attention weights across all the keypoint features is the key to our success in the heatmap head training in the LFE module.}
Compared to the \textit{reduce average} (no attention weight incorporation), the most straightforward solution to this is to use a \textit{weighted average} operation replacing the \textit{reduce average}.
We choose to use \textit{weighted average} for training as we use the FPS in the AKS module.
We choose to use \textit{reduce average} during the online localization stage and thoroughly evaluate the performance of the two different approaches in Section~\ref{subsec:ablation}.

\shiyu{The remaining part that estimates the optimal offset $\Delta \hat{z}$ and its probability distribution $P(\Delta z_i)$ for $z \in \{x, y, \psi\}$ is identical to the design of L3-Net as shown in Figure~\ref{fig:subnetwork} (c). Please refer to \cite{lu2019l3} for more details.}

\subsection{Loss}
\label{subsec:loss}


1) \textsl{Absolute Loss:} The absolute distance between the estimated offset $\Delta \hat{T}$ and the ground truth $\Delta T^* = (\Delta x^*, \Delta y^*, \Delta \psi^*)$ is applied as the first loss:
\begin{small}
\begin{equation}
\label{equ:loss1}
Loss_{1} = \alpha \cdot (| \Delta \hat{x} - \Delta x^* | + | \Delta \hat{y} - \Delta y^* | + | \Delta \hat{\psi} - \Delta \psi^* |),
\end{equation}
\end{small}
where $\alpha$ is a balancing factor.

2) \textsl{Concentration Loss:} Besides the absolute loss above, the concentration of the probability distribution $P(\Delta z_i), z \in \{x, y, \psi\}$ also has a considerable impact on the estimation robustness.
For this reason, the mean absolute deviation (MAD) of the probability distribution assuming the ground truth as the mean value is used as:
\begin{small}
\begin{equation} \label{equ:deviation}
\sigma_z = \sum_i P(\Delta z_i) \cdot | \Delta z_i - \Delta z^* |,
\end{equation}
\end{small}
where $z \in \{x, y, \psi\}$. Accordingly, the second loss function is defined as $Loss_2 = \beta \cdot (\sigma_x + \sigma_y + \sigma_\psi)$.

3) \textsl{Similarity Loss:} In addition to geometry constrains, the corresponding 2D-3D keypoint pairs should have similar descriptors. Therefore, we define the third loss:
\begin{small}
\begin{equation} \label{equ:loss3}
Loss_3 = \sum_p \max(\hat{P}(p)-C, 0),
\end{equation}
\end{small}
where $\hat{P}(p)$ is the cost volume output from the 3D CNNs of the keypoint $p$, when we project the keypoint in the map using the ground truth pose onto the online image, find the corresponding point in the online image, and compute the descriptor distance between the pair. $C = 1.0$ is a constant.

%% file: sources/exp.tex
\section{Experiments}
\label{section:exp}

\subsection{Apollo-DaoxiangLake Dataset}
\label{section:dataset}
To evaluate the proposed method, we require a dataset with multiple trials of data collection of the same road over a long period of time containing aligned camera images and 3D LiDAR scans.
To this end, we evaluated several public datasets \cite{pandey2011ford,geiger2012we,ncarlevaris2015a,maddern20171,sattler2018benchmarking,lu2019l3}.
The KITTI \cite{geiger2012we,geiger2013vision} and Ford Campus \cite{pandey2011ford} datasets fail to enclose multiple trials of the same road.
The NCLT dataset \cite{ncarlevaris2015a} is the closest one to our requirement, but unfortunately, we find that images are not well aligned with 3D LiDAR scans.
The Oxford RobotCar dataset \cite{maddern20171} doesn't provide ground truth trajectories with high confidence \shiyu{until the latest upgrade \cite{maddern2020real}.}
The Aachen Day-Night, CMU-Seasons and RobotCar-Seasons datasets \cite{sattler2018benchmarking} focus more on single-frame based localization, resulting in short trajectories lasting for only several seconds, incompatible for the application of autonomous driving.
The Apollo-SouthBay dataset \cite{lu2019l3,lu2019deepvcp} doesn't release camera images.

\shiyu{Therefore, we recruited our mapping vehicles and built this new dataset, Apollo-DaoxiangLake.
The cameras are hardware synchronized with the LiDAR sensor.
That allows us to compensate for the rolling shutter and motion effects when we project 3D point clouds onto images, yielding precise alignment between 3D point clouds and image pixels.
The ground truth poses are provided using high-end sensors through necessary post-processing solutions.}
We collected 9 trials of repetitive data within 14 weeks (Sep. to Dec.) over the same road adjacent to the Daoxiang Lake park in Beijing, China.
In particular, our dataset includes different times of the day, for example, noon, afternoon, sunset, and seasonal changes, e.g., sunny, snowy days, and difficult circumstances, e.g. foggy lens, object occlusion, making it a challenging visual localization benchmark dataset.
Some sample images are shown in Figure~\ref{fig:keypoints_seclection_visualization}.
The data from the first trial on Sep. 18 is reserved for building the localization map.
Please refer to the supplemental materials for more details about our dataset.

In our experiment, for simplicity, the input predicted poses are generated by extracting the inter-frame incremental motion from the trajectories estimated by the built-in GNSS/IMU integrated solution in NovAtel PwrPak7D-E1 with RTK disabled, and appending it to the localization output of our system at the previous frame, which is the same as the experimental setup of the LiDAR localization system in \cite{levinson2010robust}. In practice, the incremental motion should be estimated by an inertial navigation system (INS).

\subsection{Performances}
\label{subsec:performances}
\textbf{Quantitative Analysis}
\shiyu{Due to the fact that not all the methods can work properly in all circumstances in our dataset, we introduce an N/A ratio metric and allow the system to claim ``results not available'' under certain specific circumstances.
Only the results from ``available'' frames are counted in the quantitative analysis of the localization accuracy.}
For our method, we monitor the variance of estimated probability vectors $P(\Delta z_i), z \in \{x, y, \psi\}$.
We report the ``unavailable'' status when the variance is higher than a threshold.
Our quantitative analysis includes horizontal and heading (yaw) errors with both RMS and maximum values.
The horizontal errors are further decomposed to longitudinal and lateral directions.
The percentages of frames where the system achieves better than thresholds are also shown in tables.
To make a thorough evaluation, we compare the proposed approach with two methods, structure-based and feature-based.

i) Structure-based:
Following \cite{schreiber2013laneloc,jo2015precise}, a map that contains line segments to represent the lane markings or curbs is used.
2D lane segments are extracted from online images and are matched against those in the map based on a particle filter framework.
Objects, such as poles, are added to further improve the N/A ratio and longitudinal error.
Similarly, only 3 DoF are estimated in this method. When there is no adequate detected or matched lane segments, the system reports the ``unavailable'' status.

ii) Feature-based:
The latest work HF-Net \cite{sarlin2019coarse} is included.
When there is no sufficient matched inliers, the system reports the ``unavailable'' status.
The original implementation of HF-Net includes global descriptors for coarse localization (image retrieval), 6 DoF pose estimation using a PnP + RANSAC.
To conduct a fairer comparison, we made necessary modifications including three parts: 1) We replaced its global retrieval with 10 neighboring images directly found using the prior pose. 2) The local 6-DoF pose estimation using a PnP is replaced with a PnP + bundle adjustment (BA) using the 3D-2D matches from 10 to 1 images (single camera) or 30 to 3 images (multi-cameras). 3) A 3 DoF BA across $(x, y, yaw)$ dimensions is implemented.
These modifications improve its performance as shown in Table~\ref{table:dataset_performance_results}.
With regard to the feature descriptors which play an essential role during matching, we also present the experimental results using SIFT \cite{lowe2004distinctive} in the HF-Net architecture.

\renewcommand\arraystretch{1.1}
\begin{table*}[htbp]
	\centering
	\scalebox{0.73}[0.73]{
	\begin{tabular}{L{2.55cm}|R{0.6cm}R{2.1cm}R{2.1cm}R{2.1cm}R{2.1cm}R{2.1cm}R{2.1cm}} 
		\toprule[1pt]
		\multirow{2}{*}{Method}	& \multirow{2}{*}{\makecell[c]{N/A \\(\%)}} & \multicolumn{2}{c}{Horizontal} & {Longitudinal} & {Lateral} & \multicolumn{2}{c}{Yaw} \\  \cmidrule(ll){3-4} \cmidrule(ll){5-6}  \cmidrule(ll){7-8} 
		&  & {RMS/Max(m)}			& 0.1/0.2/0.3(\%) & {RMS/Max(m)} & {RMS/Max(m)} & RMS/Max(\degree)  & 0.1/0.3/0.6(\%) \\
		\midrule[0.5pt] 
	    Struct-based (S)					& 91.0 & 0.244/2.669 & 20.5/49.5/71.3 & 0.218/\textbf{1.509} & 0.076/2.533 & 0.151/4.218 & 42.1/89.7/99.0 \\
		HFNet\cite{sarlin2019coarse} (S)	& 61.4 & 0.243/322.6 & 34.3/61.4/76.3 & 0.211/322.5 & 0.081/8.844 & 0.081/15.65 & 77.8/97.8/99.6 \\
		HFNet++(S)							& 79.8 & 0.213/6.049 & 30.4/59.0/76.6 & 0.186/2.341 & 0.074/6.004 & 0.079/16.03 & 74.5/98.3/99.8 \\
		HFNet++SIFT(S)						& 41.2 & 0.264/8.181 & 28.1/54.0/70.8 & 0.211/7.349 & 0.113/8.154 & 0.106/17.42 & 75.0/96.1/98.9 \\
		HFNet++     						& 93.2 & 0.176/13.62 & 45.4/73.9/87.0 & 0.152/13.54 & 0.056/6.380 & 0.077/25.22 & 82.6/98.2/99.5 \\
		HFNet++SIFT							& 48.9 & 0.244/8.281 & 30.3/61.2/75.7 & 0.191/7.349 & 0.105/7.046 & 0.107/14.68 & 77.6/95.9/98.3 \\
		\midrule[0.5pt]
		Ours (S)							& 95.4 & 0.123/3.119 & 61.7/83.6/91.9 & 0.106/3.074 & 0.043/1.787 & 0.070/\textbf{1.685} & 80.5/97.4/99.4 \\
		Ours								& \textbf{100.0} & \textbf{0.058}/\textbf{2.617} & \textbf{86.3}/\textbf{96.8}/\textbf{99.5} & \textbf{0.048}/2.512 & \textbf{0.023}/\textbf{1.541} & \textbf{0.054}/3.208 & \textbf{89.4}/\textbf{99.6}/\textbf{99.9} \\
		\midrule[0.5pt]
		LiDAR \cite{wan2018robust}						& 100.0& 0.053/1.831 & 93.6/99.7/99.9 & 0.038/1.446 & 0.029/1.184 & 0.070/1.608 & 76.4/99.8/99.9 \\
		\bottomrule[1pt]
	\end{tabular}
	}
	\caption{
		\small
		Comparison with other methods. We achieve centimeter-level RMS errors in both longitudinal and lateral directions, which is comparable to the latest LiDAR-based solution \cite{wan2018robust}. Our overall performance, including the N/A rate and accuracy, is higher than both the structure-based and feature-based methods by a large margin.
	}
    \label{table:dataset_performance_results}
\end{table*}

In Table~\ref{table:dataset_performance_results}, we give a quantitative analysis of each method.
The method labeled ``(S)'' uses only a single front camera, others use all three cameras.
``HFNet\cite{sarlin2019coarse}'' is the original implementation with PnP +RANSAC which only works for single view. ``HFNet++'' is the method with our modifications to enhance the performance. ``HFNet++SIFT'' means that we use the SIFT descriptors in the HF-Net framework.
It demonstrates that the localization performance of our proposed vision-based method is comparable to the latest LiDAR-based solution \cite{wan2018robust}, achieving centimeter-level RMS error in both longitudinal and lateral directions.
\shiyu{We also find that the LiDAR method \cite{wan2018robust} reaches its maximum error when the road surface is wet during snowy days.}
The low localization errors of our system demonstrate that our network can generalize decently in varying lighting conditions.
The system also boosts its performance by using all the three cameras.
In addition, note our vast performance improvement over both the structure-based or feature-based methods.
The structure-based baseline method achieves high lateral precision, which is crucial for autonomous driving.
Compared to the traditional SIFT feature, the learning-based feature HF-Net demonstrates better performance.
When the global descriptors fail to locate the vehicle in the ballpark, the original HF-Net \cite{sarlin2019coarse} can produce significantly large localization errors as expected.

\textbf{Run-time Analysis}
We evaluated the runtime performance of our method with a GTX 1080 Ti GPU, Core i7-9700K CPU, and 16GB Memory. It takes 5 ms (single camera) and 14 ms (three cameras) in the preprocess step, 10 ms (single camera) and 18 ms (three cameras) in the feature embedding step, and 12 ms (single camera) and 18 ms (three cameras) in feature matching step, respectively. The total end-to-end processing time per frame during online localization is 27 ms and 50 ms for single and three cameras, respectively. The pre-built map data size is about 10 MB/km.

\subsection{Ablations and Visualization}
\label{subsec:ablation}


\textbf{Keypoint Selection and Marginalization}
We carry out a series of comprehensive experiments to compare the different keypoint selection and dimension marginalization strategies we proposed in the AKS and WFM modules in Section~\ref{section:method} with the results shown in Table~\ref{table:keypoint_selection_strategy}.
``WFPS+Weighted'' means that we choose the WFPS algorithm as our keypoint selection algorithm and the \textit{weighted average} method in our WFM module during the online localization.
Similarly, ``Reduce'' means we choose the \textit{reduce average} in the WFM module. We note that using WFPS + \textit{reduce average} outperforms others. In addition, the dramatic performance decline using FPS + \textit{reduce average} which does not incorporate the estimated attention scores, proves the effectiveness of our proposed attention mechanism.

\renewcommand\arraystretch{1.1}
\begin{table*}[h]
	\centering
	\scalebox{0.73}[0.73]{
	\begin{tabular}{L{2.55cm}|R{0.6cm}R{2.1cm}R{2.1cm}R{2.1cm}R{2.1cm}R{2.1cm}R{2.1cm}} 
			\toprule[1pt]
			\multirow{2}{*}{Method}	& \multirow{2}{*}{\makecell[c]{N/A \\(\%)}} & \multicolumn{2}{c}{Horizontal} & {Longitudinal} & {Lateral} & \multicolumn{2}{c}{Yaw} \\  \cmidrule(ll){3-4} \cmidrule(ll){5-6}  \cmidrule(ll){7-8} 
		&  & {RMS/Max(m)}			& 0.1/0.2/0.3(\%) & {RMS/Max(m)} & {RMS/Max(m)} & RMS/Max(\degree)  & 0.1/0.3/0.6(\%) \\
			\midrule[0.5pt] 
			WFPS+Weighted       & 99.9 & 0.063/\textbf{2.419} & 83.4/96.5/99.3 &  0.051/\textbf{1.811} & 0.026/2.095 &  0.056/\textbf{2.430} & 88.3/99.5/99.8 \\
			WFPS+Reduce     	& \textbf{100.0} & \textbf{0.058}/2.617 & \textbf{86.3}/\textbf{96.8}/\textbf{99.5} & \textbf{0.048}/2.512 & \textbf{0.023}/\textbf{1.541} &  \textbf{0.054}/3.208 & \textbf{89.4}/\textbf{99.6}/\textbf{99.9} \\
			FPS+Weighted       	& 98.9 & 0.306/18.40 & 58.0/76.4/82.9 &  0.222/14.96 & 0.161/17.93 &  0.195/3.390 & 66.0/87.0/92.3 \\
			FPS+Reduce       	& 98.1 & 0.135/6.585 & 69.7/85.1/90.5 &  0.109/4.643 & 0.055/6.151 &  0.105/3.287 & 76.3/93.4/97.3 \\
			\bottomrule[1pt]
		\end{tabular}	
	}
	\caption{
		\small
		Comparison with various keypoint selection and weighting strategies in our framework. WFPS + \textit{reduce average} achieves the best performance.
	}
	\label{table:keypoint_selection_strategy}
\end{table*}

\begin{figure*}[htbp]
	\centering
	\includegraphics[width=\linewidth]{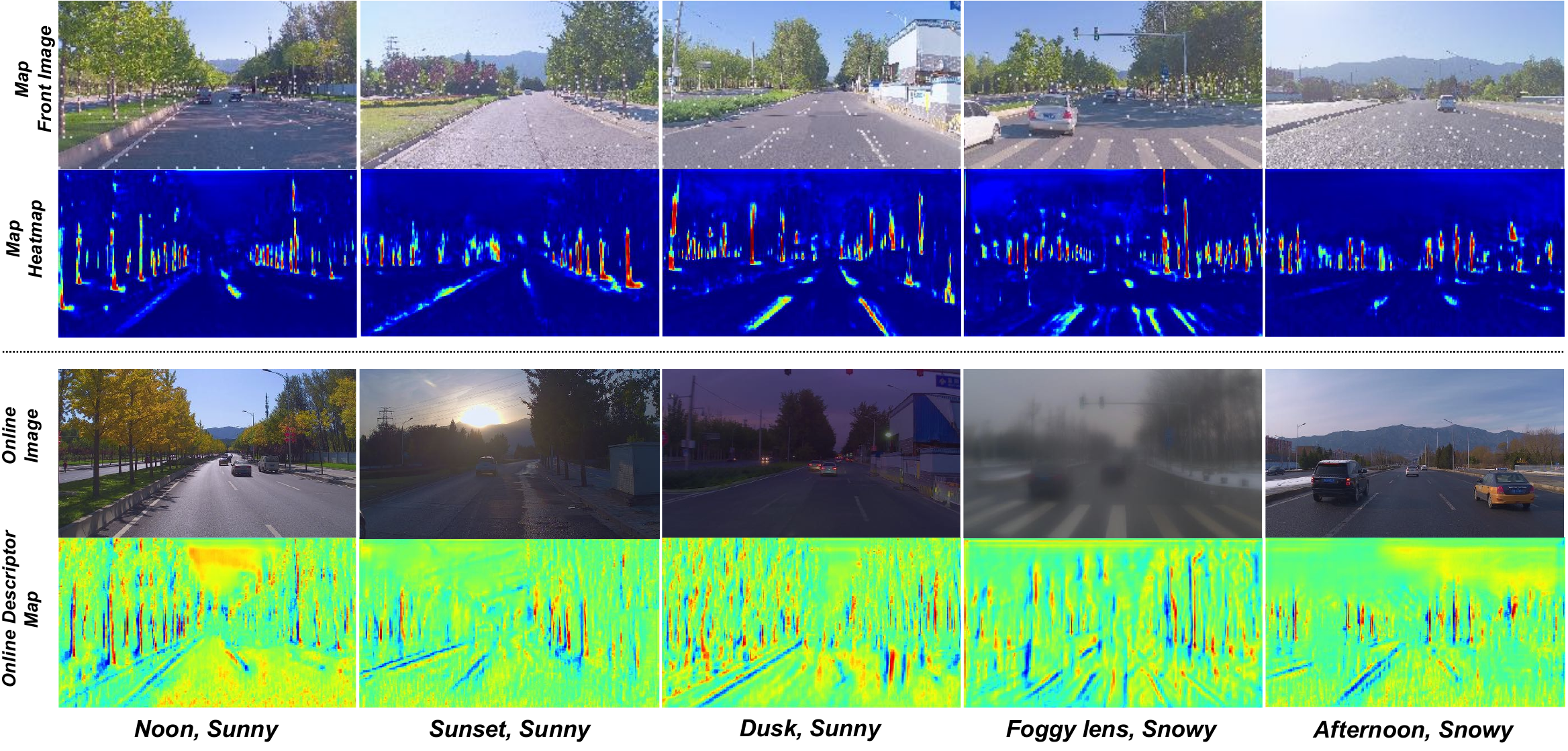}
	\caption{
		Visualization of the camera images together with the generated heatmaps, feature maps, and keypoints. Note the dramatic visual differences between the online and map images and the various challenging circumstances in our dataset.
	}
	\label{fig:keypoints_seclection_visualization}
\end{figure*}

\textbf{Keypoint and Heatmap Visualization}
To have better insights into the attention mechanism in our framework, we visualize the generated heatmaps together with the selected keypoints and the extracted feature descriptors in Figure~\ref{fig:keypoints_seclection_visualization}.
Note the diverse lighting changes from noon to sunset, dramatic seasonal changes on tree leaves and snow on the ground, challenging circumstances caused by the foggy lens.
Also interestingly, we find that the feature descriptors output by the network for dynamic objects, such as cars, are similar to the background.
It implies both our learned feature maps and heatmaps suppress the influence of the dynamic objects in our localization task.

%% file: sources/conclusion.tex
\section{Conclusion}
\label{section:conclusion}

We have presented a vision-based localization framework designed for autonomous driving applications.
We demonstrate that selecting keypoints based on an attention mechanism and learning features through an end-to-end DNN allows our system to find abundant features that are salient, distinctive and robust in the scene.
The capability of full exploitation of these robust features enables our system to achieve centimeter-level localization accuracy, which is comparable to the latest LiDAR-based methods and substantially greater than other vision-based methods in terms of both robustness and precision.
The strong performance makes our system ready to be integrated into a self-driving car, constantly providing precise localization results using low-cost sensors, thus accelerating the commercialization of self-driving cars.
\shiyu{Our future work explores building a complete vision-based localization system which may comprise both lane-based and feature-based methods, and call for the aid of odometry \cite{ziegler2014video,ding2020lio}.}

%% file: eccv2020submission.bbl
\begin{thebibliography}{10}
\providecommand{\url}[1]{\texttt{#1}}
\providecommand{\urlprefix}{URL }
\providecommand{\doi}[1]{https://doi.org/#1}

\bibitem{alahi2012freak}
Alahi, A., Ortiz, R., Vandergheynst, P.: {FREAK}: Fast retina keypoint. In:
  Proceedings of the IEEE Conference on Computer Vision and Pattern Recognition
  (CVPR). pp. 510--517. IEEE (2012)

\bibitem{barsan2018learning}
Barsan, I.A., Wang, S., Pokrovsky, A., Urtasun, R.: Learning to localize using
  a {LiDAR} intensity map. In: Proceedings of the Conference on Robot Learning
  (CoRL). pp. 605--616 (2018)

\bibitem{brahmbhatt2018geometry}
Brahmbhatt, S., Gu, J., Kim, K., Hays, J., Kautz, J.: Geometry-aware learning
  of maps for camera localization. In: Proceedings of the IEEE Conference on
  Computer Vision and Pattern Recognition (CVPR) (June 2018)

\bibitem{burki2019vizard}
B{\"u}rki, M., Schaupp, L., Dymczyk, M., Dub{\'e}, R., Cadena, C., Siegwart,
  R., Nieto, J.: {VIZARD}: Reliable visual localization for autonomous vehicles
  in urban outdoor environments. arXiv preprint arXiv:1902.04343  (2019)

\bibitem{calonder2010brief}
Calonder, M., Lepetit, V., Strecha, C., Fua, P.: {BRIEF}: Binary robust
  independent elementary features. In: Proceedings of the European Conference
  on Computer Vision (ECCV). pp. 778--792. Springer (2010)

\bibitem{ncarlevaris2015a}
Carlevaris-Bianco, N., Ushani, A.K., Eustice, R.M.: University of {Michigan}
  {North} {Campus} long-term vision and {LiDAR} dataset. International Journal
  of Robotics Research (IJRR)  \textbf{35}(9),  1023--1035 (2015)

\bibitem{caselitz2016monocular}
Caselitz, T., Steder, B., Ruhnke, M., Burgard, W.: Monocular camera
  localization in {3D} {LiDAR} maps. In: Proceedings of the IEEE/RSJ
  International Conference on Intelligent Robots and Systems (IROS). pp.
  1926--1931. IEEE (2016)

\bibitem{chen2019enforcenet}
Chen, Y., Wang, G.: {EnforceNet}: Monocular camera localization in large scale
  indoor sparse {LiDAR} point cloud. arXiv preprint arXiv:1907.07160  (2019)

\bibitem{cui2014real}
Cui, D., Xue, J., Du, S., Zheng, N.: Real-time global localization of
  intelligent road vehicles in lane-level via lane marking detection and shape
  registration. In: Proceedings of the IEEE/RSJ International Conference on
  Intelligent Robots and Systems (IROS). pp. 4958--4964. IEEE (2014)

\bibitem{cui2015real}
Cui, D., Xue, J., Zheng, N.: Real-time global localization of robotic cars in
  lane level via lane marking detection and shape registration. IEEE
  Transactions on Intelligent Transportation Systems (T-ITS)  \textbf{17}(4),
  1039--1050 (2015)

\bibitem{deTone2018superpoint}
DeTone, D., Malisiewicz, T., Rabinovich, A.: {SuperPoint}: Self-supervised
  interest point detection and description. In: Proceedings of the IEEE
  Conference on Computer Vision and Pattern Recognition Workshops (CVPRW) (June
  2018)

\bibitem{ding2020lio}
Ding, W., Hou, S., Gao, H., Wan, G., Song, S.: {LiDAR} inertial odometry aided
  robust {LiDAR} localization system in changing city scenes. In: Proceedings
  of the IEEE International Conference on Robotics and Automation (ICRA). IEEE
  (May 2020)

\bibitem{Dusmanu_2019_CVPR}
Dusmanu, M., Rocco, I., Pajdla, T., Pollefeys, M., Sivic, J., Torii, A.,
  Sattler, T.: {D2-Net}: A trainable {CNN} for joint description and detection
  of local features. In: Proceedings of the IEEE Conference on Computer Vision
  and Pattern Recognition (CVPR) (June 2019)

\bibitem{eldar1997farthest}
Eldar, Y., Lindenbaum, M., Porat, M., Zeevi, Y.Y.: The farthest point strategy
  for progressive image sampling. IEEE Transactions on Image Processing (TIP)
  \textbf{6}(9),  1305--1315 (1997)

\bibitem{fischler1981ransac}
Fischler, M.A., Bolles, R.C.: {Random Sample Consensus}: A paradigm for model
  fitting with applications to image analysis and automated cartography.
  Communications of the ACM  \textbf{24},  381--395 (1981)

\bibitem{geiger2013vision}
Geiger, A., Lenz, P., Stiller, C., Urtasun, R.: Vision meets robotics: The
  {KITTI} dataset. The International Journal of Robotics Research (IJRR)
  \textbf{32}(11),  1231--1237 (2013)

\bibitem{geiger2012we}
Geiger, A., Lenz, P., Urtasun, R.: Are we ready for autonomous driving? the
  {KITTI} vision benchmark suite. In: Proceedings of the IEEE Conference on
  Computer Vision and Pattern Recognition (CVPR). pp. 3354--3361. IEEE (2012)

\bibitem{germain2019sparse}
Germain, H., Bourmaud, G., Lepetit, V.: Sparse-to-dense hypercolumn matching
  for long-term visual localization. In: Proceedings of the International
  Conference on 3D Vision (3DV). pp. 513--523. IEEE (2019)

\bibitem{haralick1994pnp}
Haralick, B.M., Lee, C.N., Ottenberg, K., N{\"o}lle, M.: Review and analysis of
  solutions of the three point perspective pose estimation problem.
  International Journal of Computer Vision (IJCV)  \textbf{13}(3),  331--356
  (Dec 1994)

\bibitem{he2016deep}
He, K., Zhang, X., Ren, S., Sun, J.: Deep residual learning for image
  recognition. In: Proceedings of the IEEE Conference on Computer Vision and
  Pattern Recognition (CVPR). pp. 770--778 (2016)

\bibitem{he2018localdes}
He, K., Lu, Y., Sclaroff, S.: Local descriptors optimized for average
  precision. In: Proceedings of the IEEE Conference on Computer Vision and
  Pattern Recognition (CVPR) (June 2018)

\bibitem{jo2015precise}
Jo, K., Jo, Y., Suhr, J.K., Jung, H.G., Sunwoo, M.: Precise localization of an
  autonomous car based on probabilistic noise models of road surface marker
  features using multiple cameras. IEEE Transactions on Intelligent
  Transportation Systems  \textbf{16}(6),  3377--3392 (2015)

\bibitem{Cipolla2015}
Kendall, A., Grimes, M., Cipolla, R.: {PoseNet}: A convolutional network for
  real-time {6-DOF} camera relocalization. In: Proceedings of the IEEE
  International Conference on Computer Vision (ICCV). pp. 2938--2946 (Dec
  2015). \doi{10.1109/ICCV.2015.336}

\bibitem{kendall2017geometric}
Kendall, A., Cipolla, R., et~al.: Geometric loss functions for camera pose
  regression with deep learning. In: Proceedings of the IEEE Conference on
  Computer Vision and Pattern Recognition (CVPR). vol.~3, p.~8 (2017)

\bibitem{lategahn2013learn}
Lategahn, H., Beck, J., Kitt, B., Stiller, C.: How to learn an illumination
  robust image feature for place recognition. In: Proceedings of the IEEE
  Intelligent Vehicles Symposium (IV). pp. 285--291. IEEE (2013)

\bibitem{lategahn2013urban}
Lategahn, H., Schreiber, M., Ziegler, J., Stiller, C.: Urban localization with
  camera and inertial measurement unit. In: Proceedings of the IEEE Intelligent
  Vehicles Symposium (IV). pp. 719--724. IEEE (2013)

\bibitem{lategahn2014vision}
Lategahn, H., Stiller, C.: Vision only localization. IEEE Transactions on
  Intelligent Transportation Systems (T-ITS)  \textbf{15}(3),  1246--1257
  (2014)

\bibitem{Levinson_etal2007}
Levinson, J., Montemerlo, M., Thrun, S.: Map-based precision vehicle
  localization in urban environments. In: Proceedings of the Robotics: Science
  and Systems (RSS). vol.~4, p.~1 (2007)

\bibitem{levinson2010robust}
Levinson, J., Thrun, S.: Robust vehicle localization in urban environments
  using probabilistic maps. In: Proceedings of the IEEE International
  Conference on Robotics and Automation (ICRA). pp. 4372--4378 (May 2010)

\bibitem{lin2017feature}
Lin, T.Y., Doll{\'a}r, P., Girshick, R., He, K., Hariharan, B., Belongie, S.:
  Feature pyramid networks for object detection. In: Proceedings of the IEEE
  Conference on Computer Vision and Pattern Recognition (CVPR). pp. 2117--2125
  (2017)

\bibitem{linegar2015work}
Linegar, C., Churchill, W., Newman, P.: Work smart, not hard: Recalling
  relevant experiences for vast-scale but time-constrained localisation. In:
  Proceedings of the IEEE International Conference on Robotics and Automation
  (ICRA). pp. 90--97. IEEE (2015)

\bibitem{linegar2016made}
Linegar, C., Churchill, W., Newman, P.: Made to measure: Bespoke landmarks for
  24-hour, all-weather localisation with a camera. In: Proceedings of the IEEE
  International Conference on Robotics and Automation (ICRA). pp. 787--794.
  IEEE (2016)

\bibitem{lowe2004distinctive}
Lowe, D.G.: Distinctive image features from scale-invariant keypoints.
  International Journal of Computer Vision (IJCV)  \textbf{60}(2),  91--110
  (2004)

\bibitem{lu2019deepvcp}
Lu, W., Wan, G., Zhou, Y., Fu, X., Yuan, P., Song, S.: {DeepVCP}: An end-to-end
  deep neural network for point cloud registration. In: Proceedings of the IEEE
  International Conference on Computer Vision (ICCV) (October 2019)

\bibitem{lu2019l3}
Lu, W., Zhou, Y., Wan, G., Hou, S., Song, S.: {L3-Net}: Towards learning based
  {LiDAR} localization for autonomous driving. In: Proceedings of the IEEE
  Conference on Computer Vision and Pattern Recognition (CVPR). pp. 6389--6398
  (2019)

\bibitem{maddern2020real}
Maddern, W., Pascoe, G., Gadd, M., Barnes, D., Yeomans, B., Newman, P.:
  Real-time kinematic ground truth for the {Oxford} robotcar dataset. arXiv
  preprint arXiv:2002.10152  (2020)

\bibitem{maddern20171}
Maddern, W., Pascoe, G., Linegar, C., Newman, P.: 1 year, 1000 km: The {Oxford}
  {RobotCar} dataset. The International Journal of Robotics Research (IJRR)
  \textbf{36}(1),  3--15 (2017)

\bibitem{maddern2014laps}
Maddern, W., Stewart, A.D., Newman, P.: {LAPS-II}: {6-DoF} day and night visual
  localisation with prior {3D} structure for autonomous road vehicles. In:
  Proceedings of the IEEE Intelligent Vehicles Symposium (IV). pp. 330--337.
  IEEE (2014)

\bibitem{naseer2017loc}
Naseer, T., Burgard, W.: Deep regression for monocular camera-based {6-DoF}
  global localization in outdoor environments. In: Proceedings of the IEEE/RSJ
  International Conference on Intelligent Robots and Systems (IROS). pp.
  1525--1530 (Sep 2017)

\bibitem{neubert2017sampling}
Neubert, P., Schubert, S., Protzel, P.: Sampling-based methods for visual
  navigation in {3D} maps by synthesizing depth images. In: Proceedings of the
  IEEE/RSJ International Conference on Intelligent Robots and Systems (IROS).
  pp. 2492--2498. IEEE (2017)

\bibitem{noh2017large}
Noh, H., Araujo, A., Sim, J., Weyand, T., Han, B.: Large-scale image retrieval
  with attentive deep local features. In: Proceedings of the IEEE International
  Conference on Computer Vision (ICCV). pp. 3456--3465 (2017)

\bibitem{pandey2011ford}
Pandey, G., McBride, J.R., Eustice, R.M.: Ford campus vision and {LiDAR} data
  set. The International Journal of Robotics Research (IJRR)  \textbf{30}(13),
  1543--1552 (2011)

\bibitem{pascoe2015direct}
Pascoe, G., Maddern, W., Newman, P.: Direct visual localisation and calibration
  for road vehicles in changing city environments. In: Proceedings of the IEEE
  International Conference on Computer Vision Workshops (ICCVW). pp. 9--16
  (2015)

\bibitem{radwan2018vlocnet}
Radwan, N., Valada, A., Burgard, W.: {VLocNet++}: Deep multitask learning for
  semantic visual localization and odometry. IEEE Robotics and Automation
  Letters (RA-L)  \textbf{3}(4),  4407--4414 (Oct 2018)

\bibitem{ranganathan2013light}
Ranganathan, A., Ilstrup, D., Wu, T.: Light-weight localization for vehicles
  using road markings. In: Proceedings of the IEEE/RSJ International Conference
  on Intelligent Robots and Systems (IROS). pp. 921--927. IEEE (2013)

\bibitem{sarlin2019coarse}
Sarlin, P.E., Cadena, C., Siegwart, R., Dymczyk, M.: From coarse to fine:
  Robust hierarchical localization at large scale. In: Proceedings of the IEEE
  Conference on Computer Vision and Pattern Recognition (CVPR) (2019)

\bibitem{sattler2018benchmarking}
Sattler, T., Maddern, W., Toft, C., Torii, A., Hammarstrand, L., Stenborg, E.,
  Safari, D., Okutomi, M., Pollefeys, M., Sivic, J., et~al.: Benchmarking
  {6-DoF} outdoor visual localization in changing conditions. In: Proceedings
  of the IEEE Conference on Computer Vision and Pattern Recognition (CVPR). pp.
  8601--8610 (2018)

\bibitem{sattler2019aprrpr}
Sattler, T., Zhou, Q., Pollefeys, M., Leal-Taixe, L.: Understanding the
  limitations of {CNN-Based} absolute camera pose regression. In: Proceedings
  of the IEEE Conference on Computer Vision and Pattern Recognition (CVPR)
  (June 2019)

\bibitem{savinov2017quad}
Savinov, N., Seki, A., Ladicky, L., Sattler, T., Pollefeys, M.:
  {Quad-Networks}: Unsupervised learning to rank for interest point detection.
  In: Proceedings of the IEEE Conference on Computer Vision and Pattern
  Recognition (CVPR) (July 2017)

\bibitem{schreiber2013laneloc}
Schreiber, M., Kn{\"o}ppel, C., Franke, U.: {Laneloc}: Lane marking based
  localization using highly accurate maps. In: Proceedings of the IEEE
  Intelligent Vehicles Symposium (IV). pp. 449--454. IEEE (2013)

\bibitem{stewart2012laps}
Stewart, A.D., Newman, P.: {LAPS}-localisation using appearance of prior
  structure: {6-DOF} monocular camera localisation using prior pointclouds. In:
  Proceedings of the IEEE International Conference on Robotics and Automation
  (ICRA). pp. 2625--2632. IEEE (2012)

\bibitem{von2020gn}
von Stumberg, L., Wenzel, P., Khan, Q., Cremers, D.: {GN-Net}: The
  {Gauss-Newton} loss for multi-weather relocalization. IEEE Robotics and
  Automation Letters  \textbf{5}(2),  890--897 (2020)

\bibitem{suhr2016sensor}
Suhr, J.K., Jang, J., Min, D., Jung, H.G.: Sensor fusion-based low-cost vehicle
  localization system for complex urban environments. IEEE Transactions on
  Intelligent Transportation Systems (T-ITS)  \textbf{18}(5),  1078--1086
  (2016)

\bibitem{taira2018inloc}
Taira, H., Okutomi, M., Sattler, T., Cimpoi, M., Pollefeys, M., Sivic, J.,
  Pajdla, T., Torii, A.: {InLoc}: Indoor visual localization with dense
  matching and view synthesis. In: Proceedings of the IEEE Conference on
  Computer Vision and Pattern Recognition (CVPR) (June 2018)

\bibitem{toft2018semantic}
Toft, C., Stenborg, E., Hammarstrand, L., Brynte, L., Pollefeys, M., Sattler,
  T., Kahl, F.: Semantic match consistency for long-term visual localization.
  In: Proceedings of the European Conference on Computer Vision (ECCV)
  (September 2018)

\bibitem{valada2018loc}
Valada, A., Radwan, N., Burgard, W.: Deep auxiliary learning for visual
  localization and odometry. In: Proceedings of the IEEE International
  Conference on Robotics and Automation (ICRA). pp. 6939--6946 (May 2018).
  \doi{10.1109/ICRA.2018.8462979}

\bibitem{walch2017image}
Walch, F., Hazirbas, C., Leal-Taixe, L., Sattler, T., Hilsenbeck, S., Cremers,
  D.: Image-based localization using {LSTMs} for structured feature
  correlation. In: Proceedings of the IEEE International Conference on Computer
  Vision (ICCV). pp. 627--637 (2017)

\bibitem{wan2018robust}
Wan, G., Yang, X., Cai, R., Li, H., Zhou, Y., Wang, H., Song, S.: Robust and
  precise vehicle localization based on multi-sensor fusion in diverse city
  scenes. In: Proceedings of the IEEE International Conference on Robotics and
  Automation (ICRA). pp. 4670--4677 (2018)

\bibitem{wang2018dels}
Wang, P., Yang, R., Cao, B., Xu, W., Lin, Y.: {DeLS-3D}: Deep localization and
  segmentation with a {3D} semantic map. In: Proceedings of the IEEE Conference
  on Computer Vision and Pattern Recognition (CVPR) (June 2018)

\bibitem{wolcott2014visual}
Wolcott, R.W., Eustice, R.M.: Visual localization within {LiDAR} maps for
  automated urban driving. In: Proceedings of the IEEE/RSJ International
  Conference on Intelligent Robots and Systems (IROS). pp. 176--183. IEEE
  (2014)

\bibitem{wolcott2015fast}
Wolcott, R.W., Eustice, R.M.: Fast {LiDAR} localization using multiresolution
  gaussian mixture maps. In: Proceedings of the IEEE International Conference
  on Robotics and Automation (ICRA). pp. 2814--2821 (2015)

\bibitem{Wolcott_etal2017}
Wolcott, R.W., Eustice, R.M.: Robust {LiDAR} localization using multiresolution
  gaussian mixture maps for autonomous driving. The International Journal of
  Robotics Research (IJRR)  \textbf{36}(3),  292--319 (2017)

\bibitem{wu2013vehicle}
Wu, T., Ranganathan, A.: Vehicle localization using road markings. In:
  Proceedings of the IEEE Intelligent Vehicles Symposium (IV). pp. 1185--1190.
  IEEE (2013)

\bibitem{yi2016eccvlift}
Yi, K.M.Y., Trulls, E., Lepetit, V., Fua, P.: {LIFT}: Learned invariant feature
  transform. In: Proceedings of the European Conference on Computer Vision
  ({ECCV}) (2016)

\bibitem{yu2014monocular}
Yu, Y., Zhao, H., Davoine, F., Cui, J., Zha, H.: Monocular visual localization
  using road structural features. In: Proceedings of the IEEE Intelligent
  Vehicles Symposium Proceedings (IV). pp. 693--699. IEEE (2014)

\bibitem{ziegler2014video}
Ziegler, J., Lategahn, H., Schreiber, M., Keller, C.G., Kn{\"o}ppel, C., Hipp,
  J., Haueis, M., Stiller, C.: Video based localization for {BERTHA}. In:
  Proceedings of the IEEE Intelligent Vehicles Symposium (IV). pp. 1231--1238.
  IEEE (2014)

\end{thebibliography}
